%% file: main.tex
\pdfoutput=1

\documentclass[11pt]{article}

\usepackage[final]{acl}

\usepackage{times}
\usepackage{latexsym}

\usepackage{subfiles}

\usepackage{latexsym}
\usepackage{amsmath} 
\usepackage{amssymb} 
\usepackage{amsbsy} 
\usepackage{booktabs}
\usepackage{multirow}
\usepackage{pifont}
\usepackage{arydshln}

\usepackage[T1]{fontenc}

\usepackage[utf8]{inputenc}
\usepackage{todonotes}
\usepackage{microtype}

\usepackage{inconsolata}

\usepackage{graphicx}

\newcommand{\methodabbrev}{ClozeMath}

\title{ClozeMath: Improving Mathematical Reasoning in Language Models \\ by Learning to Fill Equations\thanks{This work was completed while all authors, except the fourth, were at Movian AI, Vietnam. All datasets and models were downloaded, trained, and evaluated using Movian AI's resources.}}
\author{Quang Hieu Pham$^{1\dagger}$, Thuy Duong Nguyen$^1$\thanks{The first two authors contributed equally to this work.}, Tung Pham$^1$, \\ \textbf{Anh Tuan Luu$^2$, Dat Quoc Nguyen$^1$} \\
         $^1$Qualcomm AI Research\thanks{Qualcomm Vietnam Company Limited.}; $^2$Nanyang Technological University, Singapore\\
         $^1$\texttt{\{hieupq, duonnguy, tungp, datnq\}@qti.qualcomm.com}; $^2$\texttt{anhtuan.luu@ntu.edu.sg}}

\begin{document}
\maketitle

\begin{abstract}
\subfile{Sections/Abstract.tex}
\end{abstract}

\section{Introduction}
\subfile{Sections/1_Introduction.tex}

\section{Our ClozeMath approach}\label{sec:method}

\subfile{Sections/Methods.tex}

\section{Experiments}
\subfile{Sections/Experiments.tex}

\section{Conclusion}
To better align with human-like mathematical reasoning, we introduce ClozeMath, an approach that trains language models to predict masked equations using a text-infilling objective alongside the standard language modeling objective. We highlight the pitfall of learning spurious correlations with MaskedThought, a strong current baseline, and demonstrate the consistently superior performance of ClozeMath across various tests of performance and robustness. 

\section*{Limitations}
In this work, we focus specifically on mathematical reasoning. However, we believe our approach has potential applications beyond this domain. As models are increasingly trained to leverage external tools \cite{Paranjape2023ARTAM}, they begin to exhibit human-like reasoning patterns that incorporate tool usage. Investigating the effectiveness of our method in these broader contexts would be a promising direction for future research. Due to limited computational resources, and in line with prior work such as Masked Thought \cite{chen-etal-2024-masked}, we evaluate our approach using LLMs with fewer than 10 billion parameters. Future studies could explore how our method scales when applied to larger foundational models.

\section*{Acknowledgement}
We extend our thanks to Linh The Nguyen (v.linhnt140@vinai.io) for his help in setting up the evaluation framework. 

\bibliography{custom}

\newpage 

\appendix

\label{sec:appendix}
\subfile{Sections/Appendix}

\end{document}

%% file: Sections/Abstract.tex
The capabilities of large language models (LLMs) have been enhanced by training on data that reflects human thought processes, such as the Chain-of-Thought format. However, evidence suggests that the conventional scheme of next-word prediction may not fully capture how humans learn to think. Inspired by how humans generalize mathematical reasoning, we propose a new approach named \methodabbrev\ to fine-tune LLMs for mathematical reasoning. Our \methodabbrev\ involves a text-infilling task that predicts masked equations from a given solution, analogous to cloze exercises used in human learning. Experiments on GSM8K, MATH, and GSM-Symbolic show that \methodabbrev\ surpasses the strong baseline Masked Thought in performance and robustness, with two test-time scaling decoding algorithms, Beam Search and Chain-of-Thought decoding. Additionally, we conduct an ablation study to analyze the effects of various architectural and implementation choices on our approach.

%% file: Sections/1_Introduction.tex
\begin{figure}[t!] 
    \centering
    \includegraphics[width=\columnwidth]{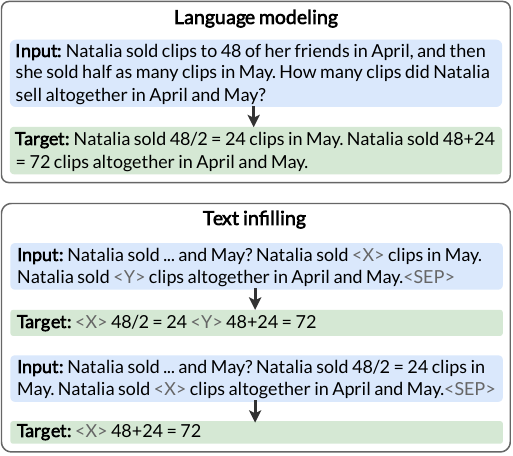} 
    \caption{\methodabbrev's mixture of training objectives. In text infilling, masked equation spans are replaced with special mask tokens (<X>, <Y>) and are gradually predicted from left to right. These mask tokens ensure alignment of token positions during training. A special <SEP> token separates the input and target parts, using bidirectional and causal attention masks, respectively.}
    \label{fig:UL-Math-demo}
\end{figure}

To mimic human reasoning in mathematical settings, current Large Language Models (LLMs) are prompted or trained to generate intermediate thinking steps before reaching a conclusion \cite{wei2022chain, kojima2022large}. Some efforts focus on generating data in this format by collecting responses from humans or LLMs \cite{lightman2023let, yue2024mammoth} or by implicitly learning task-specific representations \cite{wang2024guiding}.   However, while these outputs resemble the surface form of human thought, the current language modeling objective may not align with how humans learn to think \cite{pmlr-v235-bachmann24a, stochasticparrot}. When learning from a mathematical solution, it is easier for a human to generalize the main approach before working on the details, rather than memorizing which step follows another \cite{mason2010thinking}. Considering the input problem in Figure \ref{fig:UL-Math-demo}, the textual part of the target solution describes its general approach (i.e., rationales), while the equations are problem-specific.

In light of this observation, we propose enhancing the training of language models for mathematical reasoning by introducing an additional text infilling objective \cite{2020t5,tay2023transcending, UL2}. In this approach, models learn to predict masked equations alongside the standard language modeling objective. This method reinforces the model's ability to infer mathematical relationships by predicting equations based on the surrounding rationale, encouraging a more structured understanding of problem-solving steps.

While traditional text infilling objectives involve masking and predicting random text spans, we find that this approach disrupts the logical coherence of the unmasked parts available to the model, ultimately harming reasoning performance. Regarding model architecture, PrefixLM \cite{j.2018generating}, which applies a bidirectional attention mask to the model's prompt, has been shown to work well with text infilling in T5 \cite{2020t5}. However, in the context of mathematical reasoning, we demonstrate that only our equation masking strategy benefits from this architectural choice. 

From a different perspective, Masked Thought \cite{chen-etal-2024-masked}, a recent method with strong empirical results, regularizes a model to look further back into a math problem's definition by randomly corrupting the problem's solution during language model training. However, mathematical solutions may include auxiliary derived steps that are not directly connected to the problem definition. We demonstrate a common example in which Masked Thought may teach the model spurious correlations in Section \ref{sec:main_results}. Empirically, our method also substantially outperforms Masked Thought.

In summary, our contributions are as follows: \textbf{(1)} We introduce a simple yet highly effective fine-tuning strategy, namely ClozeMath, that enhances the mathematical reasoning abilities of LLMs. \textbf{(2)} We conduct extensive experiments on two standard benchmarks, GSM8K \cite{cobbe2021training} and MATH \cite{hendrycks2021measuring}, including a study on \methodabbrev\ scalability on test-time compute, and perform robustness analyses on the newly released benchmark GSM-Symbolic \cite{mirzadeh2024gsmsymbolic}. We also show that ClozeMath outperforms the recent strong baseline Masked Thought. \textbf{(3)} We present a comprehensive ablation study that explores the impact of various architectural choices on the effectiveness of our proposed method.

%% file: Sections/Methods.tex
Let $\{X^i, Y^i\}_{i=1}^n$ be a training set, in which $X^i$ is a text-based mathematical problem (e.g., $X^i$ is the Input in the Language modeling part in Figure \ref{fig:UL-Math-demo}), and $Y^i = y^i_1\ y^i_2\ \dots\ y^i_{m_i}$, where each $y^i_m$ is a text segment or a mathematical equation (e.g., $Y^i$ is the Target in the Language modeling part in Figure \ref{fig:UL-Math-demo}; then $m_i=5$ and $y^i_1$ is "Natalia sold", $y^i_2$ is "48/2 = 24", $y^i_3$ is "clips in May. Natalia sold", $y^i_4$ is "48+24 = 72" and $y^i_5$ is "clips altogether in April and May.").  
The language modeling objective is formulated as:
\begin{align}
    \mathcal{L}_{\mathrm{lm}} = &\sum_{i=1}^n P(Y^i|X^i)
    \label{eq:loss_lm}
\end{align}

Let $F^i = \{f_1^i, f^i_2,\dots\,f^i_{|F^i|}\} \subset \{1, 2,..., m_i\}$ be the set of indices such that $\forall m \in F^i$, $y_m^i$ is a mathematical equation (e.g., from the example in the paragraph above, $F^i = \{2, 4\}$). Our text-infilling objective is formulated as follows:

{\small
\begin{align}
    \label{eq:loss_tf}
    \mathcal{L}_{\mathrm{tf}} = &\sum_{i=1}^n \sum_{\mathrm{M} \in \mathcal{M}(F^i)} P(y^i_{m \in \mathrm{M}} | X^i, Y^i\setminus y^i_{m \in \mathrm{M}})
\end{align}
}

\noindent where $\mathcal{M}(F^i) = \{\{f^i_{t:|F^i|}\} \mid t=1, \dots, |F^i|\}$ contains combinations of equations to be masked (e.g., given $F^i = \{2, 4\}$, we have $\mathcal{M}(F^i) = \{\{2, 4\}, \{4\}\}$). This involves masking all equations and then gradually unmasking them from the first to the last, as each equation only involves computations from its previous counterparts. For groups of simultaneous mathematical transformations, such as systems of equations, we randomly mask 50\% of each group.

Finally, our ClozeMath approach trains to minimize the two objectives simultaneously:
\begin{align}
    \mathcal{L}_{\text{\methodabbrev}} = \mathcal{L}_{\mathrm{lm}} + \mathcal{L}_{\mathrm{tf}}
\end{align}
\noindent Since the number of terms in $\mathcal{L}_{\mathrm{tf}}$ depends on the number of equations to be masked within each dataset, in practice, we balance the two training objectives by duplicating samples optimized for $\mathcal{L}_{\mathrm{lm}}$ to maintain a roughly 50:50 sample ratio.

We implement ClozeMath with PrefixLM \cite{j.2018generating} on pre-trained decoder-only models. In PrefixLM, the prompt part's attention is bidirectional, while the target sequence uses a causal mask. To let the model learn to use the two attention patterns correctly, there is a special \textit{<SEP>} token separating the two parts. 
During inference, we employ the conventional next-token prediction.

%% file: Sections/Experiments.tex
\begin{figure*}[t!] 
    \centering
    \includegraphics[width=\linewidth]{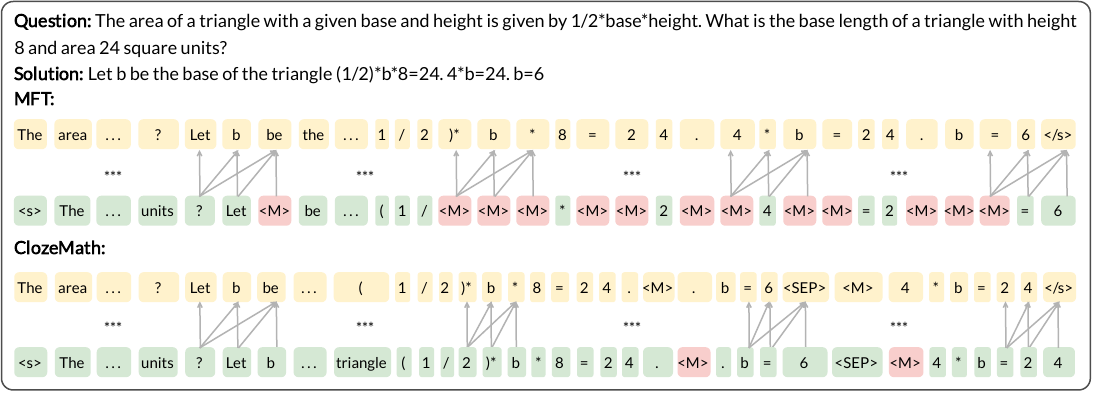} 
\caption{Implementation of MFT \cite{chen-etal-2024-masked} and \methodabbrev\ on the same sample. Here, *** denotes the unshown causal attention mask in this figure. MFT aims at learning longer-range dependencies by randomly inputting noisy mask (<M>) tokens into the solution's parts while keeping the problem definition untouched. MFT is prone to learning spurious correlation; in this example, a derived transformation might not fully perceive its predecessors.}
    \label{fig:MFT_demo}
\end{figure*}

\subsection{Experimental setup}

\textbf{Datasets:}\ \ We conduct experiments on two standard benchmarks for mathematical problem solving, GSM8K \cite{cobbe2021training} and MATH \cite{hendrycks2021measuring}.  To demonstrate the context-understanding ability of our method, we also include the GSM-Symbolic evaluation collection \cite{mirzadeh2024gsmsymbolic}, which varies problems from the GSM8K test set by: (1) changing names, nouns, and numbers (GSM-Sym) (2) adding a new constraint (GSM-P1), and (3) adding two new constraints (GSM-P2), as additional evaluation sets. See Appendix \ref{appendix:datasets} for more details. 

\noindent \textbf{Implementation:} To verify \methodabbrev\ under a consistent fine-tuning condition, instead of continually fine-tune instruction following models, we fine-tune recent strong base LLMs, i.e models without the instruction following ability, including three generic models "Llama-3.1-8B", "Llama-3.2-3B", "Llama-3.2-1B" \cite{grattafiori2024llama3herdmodels}, and a mathematics-specific model "DeepSeek-Math-7B-base" \cite{shao2024deepseekmath}, with low-rank adaptation LoRA (rank=32) \cite{hu2022lora}; and other hyperparameters are specified in Apppendix \ref{appendix:train&inference config}. Without specification, we report results of all our models and baselines obtained by \textbf{beam search} decoding with num\_beams = 5.

Here, we also extend the foundation models' vocabulary and fine-tune their available reserved tokens for \textit{<SEP>} and mask tokens.

\noindent \textbf{Baseline:}\ \ Our strong baseline approach for mathematical reasoning is  Masked Thought Fine-tuning \cite{chen-etal-2024-masked} using the same LoRA setting.

\subsection{Main results}
\label{sec:main_results}

\paragraph{Comparison with Masked Thought Fine-tuning (MFT):} Figure \ref{fig:MFT_demo} demonstrates the MFT approach. When learning from a math problem, MFT randomly masks solution tokens while preserving their corresponding labels and the problem definition unmasked. This disruption to the solution's logical flow forces the model to generate tokens by focusing on previous unmasked information, where the problem definition serves as the largest consecutive informative segment. In the example in Figure \ref{fig:MFT_demo}, the solution's first step formulates all information from the problem definition, with later steps being direct transformations of this initial step. As a result, MFT's masking strategy might learn spurious correlations when transformation steps are closely interconnected without direct links to the problem definition. For instance, the model is forced to predict "$4*b = 24$" while the definition of variable "$b$" is masked in the prior context.

\begin{table}[t!]
\centering
\setlength{\tabcolsep}{0.2em}
\resizebox{\columnwidth}{!}{
\begin{tabular}{llcccc}
\toprule
                       & \textbf{Settings}       & \textbf{DeepSeek-Math}     & \textbf{LLama-3.1-8B}    & \textbf{LLama-3.2-3B} & \textbf{LLama-3.2-1B} \\ \hline  
\multirow{3}{*}{\rotatebox[origin=c]{90}{GSM8K}} 
                       & Base           & 59.21             & 49.58         & 17.66 & 4.62 \\
                       & MFT            & 70.20             & 64.82    & 45.03   & 21.15    \\
                       & \methodabbrev  & \textbf{74.22}             & \textbf{70.00}              & \textbf{53.15}   & \textbf{27.89} \\ \midrule
\multirow{3}{*}{\rotatebox[origin=c]{90}{MATH}}
                       & Base           &  31.68                & 18.06    & 4.54 & 3.84  \\
                       & MFT            &  33.42                              & 20.94     & 10.14 & 4.52  \\
                       & \methodabbrev  & \textbf{36.90}           & \textbf{22.88}       & \textbf{11.18} & \textbf{5.00} \\
                       \bottomrule        
\end{tabular}
}
\caption{Overall results  on GSM8K and MATH datasets. "DeepSeek-Math" and "MFT" stand for "DeepSeek-Math-7B-base" and the baseline Masked Thought Fine-tuning, respectively. "Base" denotes the results obtained with few-shot prompting (5-shot for GSM8K and 4-shot for MATH) on the base models (without fine-tuning) using the well-known "lm-evaluation-harness" framework \cite{eval-harness}.}
\label{tab:overall}
\end{table}

\begin{figure}[t!]
    \centering
    \includegraphics[width=\columnwidth]{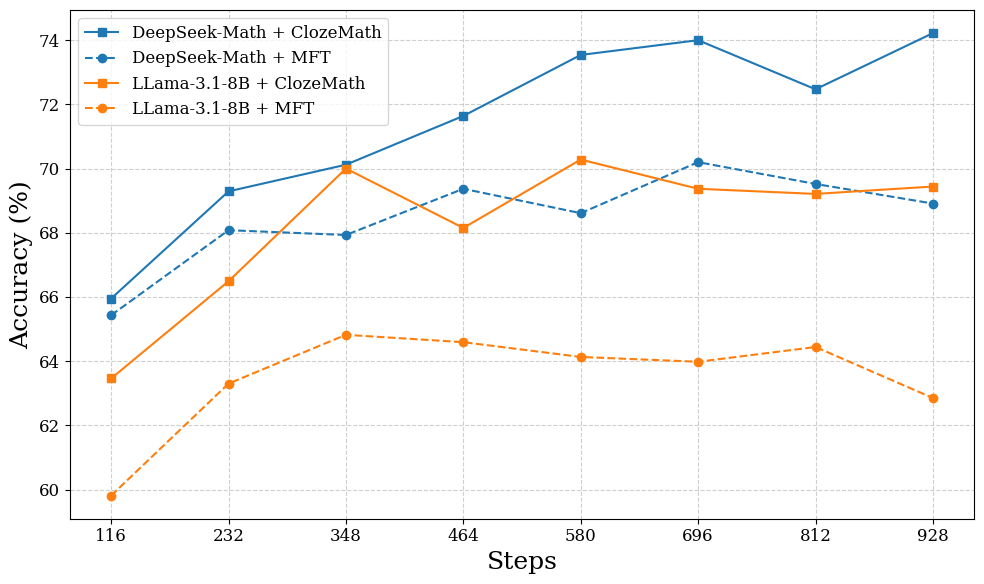} 
    \caption{GSM8K test accuracies w.r.t. the numbers of training steps.}
    \label{fig:acc_through_steps}
\end{figure}

In contrast, \methodabbrev's equation text-infilling strategy provides the model with a general solution plan and teaches it the correlation between consecutive transformation steps, which is more reasonable. Empirically, Table \ref{tab:overall} demonstrates that \methodabbrev \ substantially outperforms MFT on both GSM8K and MATH datasets. Furthermore, Figure \ref{fig:acc_through_steps} shows that \methodabbrev \ is also more sample efficient than MFT, consistently showing better performance across all recorded training checkpoints. We put some qualitative examples in Appendix \ref{appendix:qualitative_examples}.

\begin{table}[t!]
\centering
\begin{tabular}{lcc}
\toprule
    \textbf{Model}      & \textbf{MFT} & \textbf{\methodabbrev} \\ \hline      
    DeepSeek-Math   & 76.50     & \textbf{77.10} \\
    LLama-3.1-8B    & 74.30     & \textbf{75.97} \\
    Llama-3.2-3B    & 55.27     & \textbf{57.39} \\
    Llama-3.2-1B    & 28.80     & \textbf{31.84} \\ \bottomrule        
\end{tabular}
\caption{Experimental results on GSM8K with Chain-of-Thought decoding, a type of self-consistency decoding with the number of generated samples = 9 \cite{wang2024chainofthought}.}
\label{tab:cot-decoding}
\end{table}

\paragraph{Model's generation to test-time scaling:} Besides sample efficiency during training, the ability to leverage more computational budget during inference time has been proven to be an important aspect that contributes to the success of recent reasoning models \cite{gpto1, snell2025scaling}. To prove the efficiency of our method during inference time, we implement a type of self-consistency decoding method, namely Chain-of-thought (CoT) decoding \cite{wang2024chainofthought}. For each test example, CoT decoding samples $k$ answers starting with the first $k$ most probable tokens. CoT decoding then computes and aggregates by summing sampled answers' scores based on the model's confidence in answer tokens. Specifically, on GSM8K, answer tokens are numerical tokens after the annotation string \textit{\#\#\#\#}. The answer with the highest aggregated score is the model's final answer. However, due to inconsistencies in models’ tokenizer decoding and encoding processes, where a sequence of token IDs may be re-encoded differently after decoding, the default output format of the MATH dataset poses a compatibility issue. Specifically, numerical answers in MATH are wrapped inside \texttt{\textbackslash boxed\{\}}, which may not always appear at the end of the output sequence. This introduces a risk of incorrect alignment for scoring during CoT decoding, making the format incompatible with the method's confidence score computations. Consequently, we limit our evaluation  on GSM8K only, with $k=9$ sampled answers per question. As shown in Table \ref{tab:cot-decoding}, the results consistently demonstrate that \methodabbrev\ achieves better scaling than MFT with the same inference-time computation budget.

\paragraph{Models' robustness on GSM-Symbolic:} 
We further evaluate the models fine-tuned using GSM8K and MATH with MFT and ClozeMath on 5K GSM-Symbolic examples and present the results in Table \ref{tab:gsm_symbolic}. Overall, \methodabbrev\ notably outperforms MFT across different evaluation sets. For DeepSeek-Math and Llama-3.1-8B trained on GSM8K, \methodabbrev\ achieves a 5.0\% improvement (49.25\% vs. 44.25\%) and a 6.35\% improvement (47.65\% vs. 41.30\%) over MFT on GSM-P1, where an added constraint shifts the problem distribution from GSM8K. Similarly, the same trends are observed with Llama-3.2-3B and Llama-3.2-1B, indicating that \methodabbrev\ generalizes well across different model scales. On GSM-P2, which is considerably harder than GSM8K, the models fine-tuned with both methods on GSM8K show only a slight performance gap. However, when trained on MATH, which is less similar to GSM-Symbolic than GSM8K, the gap between \methodabbrev\ and MFT becomes clearer on GSM-P2: 2.2\% (16.9\% vs. 14.7\%) for DeepSeek-Math and 1.0\% (8.5\% vs. 7.5\%) for Llama-3.1-8B. We hypothesize that the small performance gap between \methodabbrev \ and MFT on Llama-3.2-1B, trained on the MATH dataset, is due to its high complexity and broad domain coverage, making it challenging for a 1B model to learn effectively.

\begin{table*}[t!]
\resizebox{2\columnwidth}{!}{
\begin{tabular}{lllcccc}
\toprule
Train. & Evaluation                 & Settings          & DeepSeek-Math                  & Llama-3.1-8B   & Llama-3.2-3B & Llama-3.2-1B       \\          \midrule
\multirow{8}{*}{\rotatebox[origin=c]{90}{\textbf{GSM8K} training set}}  & \multirow{2}{*}{GSM-Sym}              
                          & MFT                 & 63.05                          & 60.90 & 40.45 & 18.55 \\               
&                         & \methodabbrev     & \textbf{65.65}                          & \textbf{63.40}     & \textbf{46.70} & \textbf{23.95} \\ \cdashline{2-7}
& \multirow{2}{*}{GSM-P1}  & MFT                 & 44.25                          & 41.30                    & 22.5 & 8.30 \\              
&                          & \methodabbrev     & \textbf{49.25}                          & \textbf{47.65}    & \textbf{28.6} & \textbf{11.25} \\ \cdashline{2-7}
& \multirow{2}{*}{GSM-P2}  & MFT                 & \textbf{20.20}                          & 22.30           & 6.80 & 1.30 \\     
 &                         & \methodabbrev     & 20.00                          & \textbf{22.50}             & \textbf{10.20} & \textbf{3.20} \\ \cline{2-7}
& \multirow{2}{*}{Overall} & MFT                 & 46.96                          & 45.34                    & 26.54 & 11.00 \\            
&                          & \methodabbrev     & \textbf{50.08}                          & \textbf{48.92}    & \textbf{32.16} & \textbf{14.72} \\ \bottomrule     
\multirow{8}{*}{\rotatebox[origin=c]{90}{\textbf{MATH} training set}}  & \multirow{2}{*}{GSM-Sym}  
                          & MFT             &   55.50                           &  39.85                     & 17.40 & 4.10\\
&                         & \methodabbrev   &   \textbf{58.35}                           &  \textbf{45.85}   & \textbf{25.10} & 4.05\\ \cdashline{2-7}
& \multirow{2}{*}{GSM-P1}  & MFT            &   35.00                           &  21.80                     & 7.85 & 1.40\\
&                          & \methodabbrev  &   \textbf{40.25}                      &  \textbf{23.90}        & \textbf{8.45} & \textbf{1.60} \\ \cdashline{2-7}
& \multirow{2}{*}{GSM-P2}  & MFT            &   14.70                           &  7.50                      & 2.30 & 1.00 \\
 &                         & \methodabbrev  &   \textbf{16.90}                           &  \textbf{8.50}    & 1.80 & 1.00 \\ \cline{2-7}
& \multirow{2}{*}{Overall} & MFT            &   39.14                           &  26.16                     & 10.56 &  2.40\\
&                          & \methodabbrev  &   \textbf{42.82}                      &  \textbf{29.60}        & \textbf{13.78} & \textbf{2.46}\\ \bottomrule  
\end{tabular}
}
\caption{Results on GSM-Symbolic evaluation sets for the fine-tuned models used to report scores in Table \ref{tab:overall}.}
\label{tab:gsm_symbolic}
\end{table*}

\begin{table}[t!]
\centering 
\resizebox{\columnwidth}{!}{
\begin{tabular}{lc}
\toprule
\multicolumn{1}{l}{Model}                                                        &  \multicolumn{1}{c}{GSM8K} \\ \midrule
ClozeMath\textsubscript{DeepSeek-Math} & \textbf{74.22} \\
\hdashline
\ \ \ \ W/o Text infilling & 71.79 \\
\ \ \ \ W/o PrefixLM & 72.71 \\
\ \ \ \ W/o Text-infilling \& W/o PrefixLM  &  71.57 \\
\ \ \ \ W/o Equation masking & 71.19 \\
\hline 
\end{tabular}
}
\caption{Ablation study results.}
\label{tab:ablationresults}
\end{table}

\subsection{Ablation study}
\label{sec:ablation}
We examine the effectiveness of our equation masking strategy and architectural choices, as described in Section \ref{sec:method}, using DeepSeek-Math-7B on GSM8K, and present obtained results in Table \ref{tab:ablationresults}.

\textbf{W/o Text-infilling} (71.79\%) refers to the variant optimized with only the language modeling objective $\mathcal{L_{\mathrm{lm}}}$ using PrefixLM, without the text filling objective. Table \ref{tab:ablationresults} shows the notable contribution of our text infilling objective (i.e., improving the performance from 71.79\% to 74.22\%). 
\textbf{W/o PrefixLM} (72.71\%) refers to the variant that employs causal language modeling (CausalLM) instead of PrefixLM.
Therefore, \textbf{W/o PrefixLM \& W/o Text-infilling} (71.57\%) represents the conventional \textbf{instruction tuning} setting (\textbf{IT}), where only $\mathcal{L_{\mathrm{lm}}}$ is optimized with CausalLM. 

\citet{2020t5} point out that a causal attention mask limits the model's state representation because the prefix, such as a problem definition, is always included in the model's context. However, when trained exclusively with IT, PrefixLM only slightly improves the model's accuracy from 71.57\% to 71.79\%. This suggests that the model struggles to comprehend problem definitions.
In contrast, when combined with our equation text-infilling objective, PrefixLM substantially boosts the model's performance, surpassing the standard IT setup by 2.65\% (71.57\% $\rightarrow$ 74.22\%). Note that without PrefixLM, the result obtained with our equation infilling objective is still higher than that of the IT setting by 1.14\% (72.71\% vs. 71.57\%). This confirms that the presence of textual rationales alongside the problem definition in the model's prefix during training makes learning to fill equations easier. As a result, the model learns to fill in equations more effectively and develops a deeper and more structured understanding of each problem.

\textbf{W/o Equation masking:} \ \ 
To support our claim, we compare our masking strategy to a setting where the text infilling objective is optimized over randomly masked spans instead of mathematical equations. Specifically, we randomly mask both short and long token spans in each example. Depending on the span length used, we mask out either 15\% (for short spans) or 50\% (for long spans) of each solution. This approach mimics the equation masking strategy by considering the number of masked tokens in each solution. Table \ref{tab:ablationresults} shows that random masking disrupts the logical meaning of the textual parts of mathematical solutions, substantially reducing the model's performance from 74.22\% to 71.19\%.

%% file: Sections/Appendix.tex
\section{Datasets}\label{appendix:datasets}

\paragraph{GSM8K:} The dataset contains $8,752$ middle school math problems ($7,473$ training, $1,319$ test). In general, problems in GSM8K involve using only basic arithmetic operations ($+, -, \times, \div$). The dataset's \textit{main} configuration features step-by-step textual solutions with annotations of calculations provided in each solution's reasoning step.
\paragraph{MATH:} The dataset consists of $12,500$ problems ($7,500$ training, $5,000$ test) and is significantly more challenging than GSM8K, covering topics like calculus, geometry, probability, and number theory. Its solutions include equations with \LaTeX{} annotations and Asymptote vector graphics.
\paragraph{GSM-Symbolic:} The dataset is constructed from $100$ templates based on GSM8K's test problems by introducing common names, nouns, and numbers in both problems and solutions as symbolic variables to generate a diverse set of problems. The initial release includes three difficulty levels: (i) GSM-Sym, which only replaces variables in the original problems; (ii) GSM-P1, with a new constraint added to the problems before replacing variables; and similarly (iii) GSM-P2, with two new constraints. To balance cost and diversity, we sample $5000$ problems from the GSM-Symbolic dataset for evaluation.

\section{Training \& Inference configurations} 
\label{appendix:train&inference config}
We train all of our models and the baseline with LoRA adapters (rank=32), learning\_rate=5e-5, and a cosine learning rate scheduler, on a machine with a single 40GB NVIDIA A100 GPU. 

\section{Qualitative examples}
\label{appendix:qualitative_examples}

\begin{figure*}[h!]
    \centering
    \includegraphics[width=\textwidth]{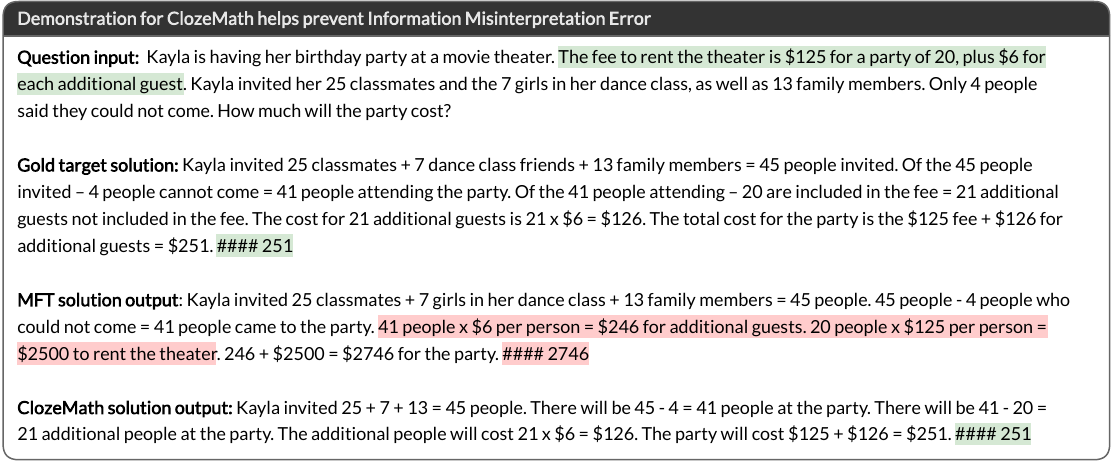} 
    \caption{In this example, the Deepseek-Math model trained with MFT does not understand that the $\$6$ prices are only applied for the number of people exceeding 20 (highlighted in green in the "Question input:").}
    \label{fig:formula_confusion_error}
\end{figure*}

\begin{figure*}[h!]
    \centering
    \includegraphics[width=\textwidth]{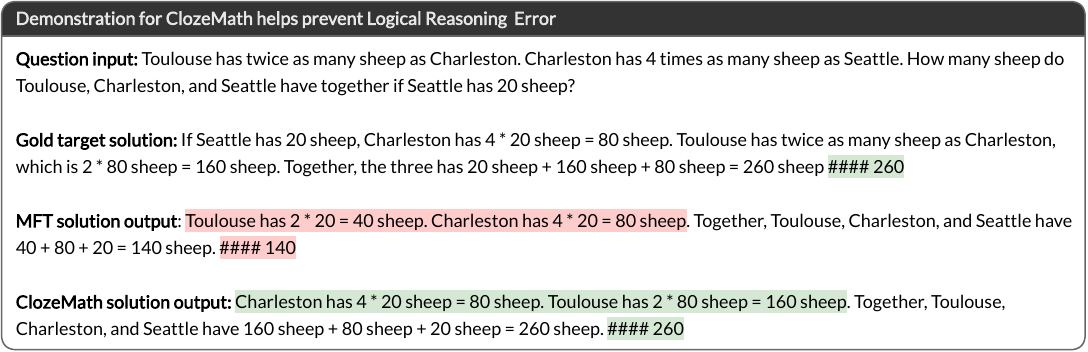} 
    \caption{In this example, the Deepseek-Math model trained with MFT misunderstood that "Toulouse" has twice the number of sheep as "Seattle," while the double relationship is actually between Toulouse's number of sheep and Charleston's number of sheep.}
    \label{fig:logic_error}
\end{figure*}

\begin{figure*}[h!]
    \centering
    \includegraphics[width=\textwidth]{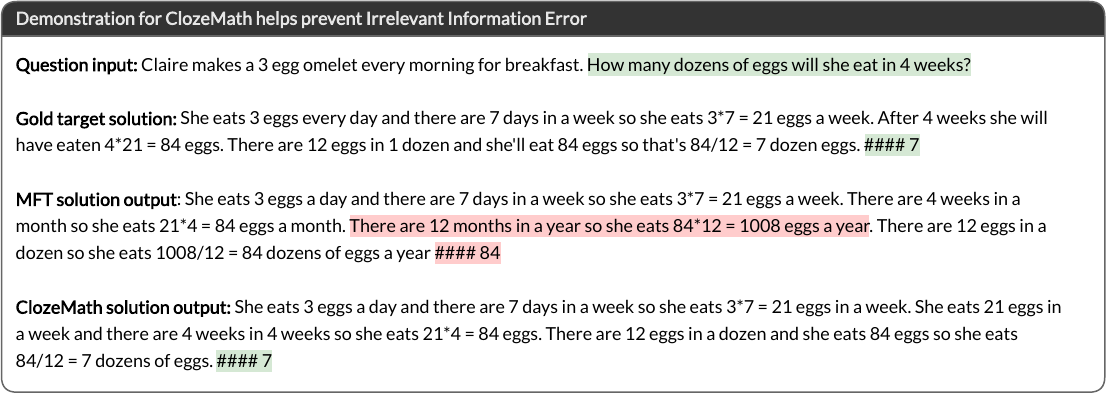} 
    \caption{In this example, the Deepseek-Math model trained with MFT incorrectly hallucinates that "There are 12 months in a year..." even though the question does not ask for this information, leading to an incorrect final calculation.}
    \label{fig:hallucinate_error}
\end{figure*}

We select some representative examples to demonstrate cases where \methodabbrev\ performs better than MFT, such as a better understanding of numeric values, as illustrated in Figures \ref{fig:formula_confusion_error} and \ref{fig:logic_error}, and the prevention of hallucination, as shown in Figure \ref{fig:hallucinate_error}.

